%% file: main.tex
\documentclass[sigconf, 10pt]{acmart}

\usepackage{array}
\usepackage{booktabs} 
\usepackage{enumerate}
\usepackage{stfloats}
\usepackage{footmisc}

\usepackage{graphicx}
\usepackage{caption}
\usepackage{subcaption} 
\usepackage{enumitem}
\usepackage{diagbox}
\usepackage{bbm}
\usepackage{bm}
\usepackage{makecell}

\usepackage{arydshln}
\usepackage{multirow}
\usepackage{amsmath}
\usepackage[noend]{algpseudocode}
\usepackage{algorithm}
\usepackage{color}
\usepackage[export]{adjustbox}
\newcommand{\xl}[1]{\textcolor{teal}{[LL: #1]}}

\usepackage{bbding}

\newcommand{\workname}{Llambda}

\usepackage{tikz}

\acmConference[Preprint]{Preprint}{TBD}{TBD}

\begin{document}
    \title[\workname]{An LLM-Empowered Low-Resolution Vision System for On-Device Human Behavior Understanding}
    \author{Siyang Jiang$^{\dagger,1}$, Bufang Yang$^{\dagger,1}$, Lilin Xu$^{2}$, Mu Yuan$^{1}$, Yeerzhati Abudunuer$^{1}$, Kaiwei Liu$^{1}$, Liekang, Zeng$^{1}$, Hongkai Chen$^{1}$, Zhenyu Yan$^{1}$, Xiaofan Jiang$^{2}$, Guoliang Xing$^{1}$}
    \affiliation{%
    \institution{$^{1}$ The Chinese University of Hong Kong, Hong Kong SAR \\
    $^{2}$ Columbia University, United States}
    }
    \thanks{$^{\dagger}$ denotes equal contribution. Guoling Xing is the corresponding author.}
   \begin{abstract} The rapid advancements in Large Vision Language Models (LVLMs) offer the potential to surpass conventional labeling by generating richer, more detailed descriptions of on-device human behavior understanding (HBU) in low-resolution vision systems, such as depth, thermal, and infrared.
   However, existing large vision language model (LVLM) approaches are unable to understand low-resolution data well as they are primarily designed for high-resolution data, such as RGB images. A quick fixing approach is to caption a large amount of low-resolution data, but it requires a significant amount of labor-intensive annotation efforts.
   
   In this paper, we propose a novel, labor-saving system, \workname, designed to support low-resolution HBU. The core idea is to leverage limited labeled data and a large amount of unlabeled data to guide LLMs in generating informative captions, which can be combined with raw data to effectively fine-tune LVLM models for understanding low-resolution videos in HBU. First, we propose a \textit{Contrastive-Oriented Data Labeler}, which can capture behavior-relevant information from long, low-resolution videos and generate high-quality pseudo labels for unlabeled data via contrastive learning. Second, we propose a \textit{Physical-Knowledge Guided Captioner}, which utilizes spatial and temporal consistency checks to mitigate errors in pseudo labels. Therefore, it can improve LLMs' understanding of sequential data and then generate high-quality video captions. Finally, to ensure on-device deployability, we employ \textit{LoRA-based efficient fine-tuning} to adapt LVLMs for low-resolution data. We evaluate \workname~using a region-scale real-world testbed and three distinct low-resolution datasets, and the experiments show that  \workname~outperforms several state-of-the-art LVLM systems up to $40.03\%$ on average Bert-Score.  
    \end{abstract}



    \settopmatter{printacmref=false}
    \renewcommand{\footnotetextcopyrightpermission}[1]{}

    \settopmatter{printacmref=false}      
    \settopmatter{printfolios=true}       
    \renewcommand\shortauthors{Siyang Jiang, Bufang Yang et al.}  
    \renewcommand\shorttitle{\workname}  

    \maketitle

\input{1_intro}

    \input{3_application}
    \input{4_overview}

    \input{5_design}
    \input{6_exp}
    \input{2_related}

\input{8_con}

    \bibliographystyle{ACM-Reference-Format}
    \bibliography{main}
\end{document}

%% file: 1_intro.tex
\section{Introduction} 
\label{sec:intro}



Compared to the conventional human action recognition (HAR) tasks~\cite{sun2022human,ouyang2022cosmo}, Human Behavior Understanding (HBU) aims to generate descriptive sentences from video recordings of human activities~\cite{li2021uav}, with a broad range of applications spanning Alzheimer's disease prevention and intervention~\cite{ouyang2024admarker,arai2021influence}, daily living monitoring~\cite{ouyang2022cosmo,jiang2024artfl}, and in-home clinical guidance~\cite{xie2021vitalhub,yang2024drhouse}.
In real-world HBU applications, low-resolution sensors (e.g., depth, thermal, and infrared) are often prioritized over high-resolution RGB data due to their reduced sensitivity to personal identifiers while maintaining sufficient behavioral information. A typical requirement of such HBU systems is to generate detailed and informative captions from low-resolution videos. 


Large Vision Language Models (LVLMs) offer promising potential for meeting this requirement.\footnote{The key difference between Large Vision Language Models (LVLMs) and Large Language Models (LLMs) in this work is that LVLMs possess the capability for video understanding.}
While recent LVLMs like Qwen-VL~\cite{Qwen2.5-VL} and OneLLM~\cite{han2023onellm} demonstrate exceptional multimodal understanding, their direct application to low-resolution data analysis remains constrained by their training paradigm.
These models primarily learn from high-resolution ``⟨RGB, caption⟩'' pairs in general vision-language tasks, which struggle to perform well on low-resolution HBU tasks.
A straightforward solution is to collect and label abundant ``⟨low-resolution data, caption⟩'' pairs for training specialized LVLMs. However, this approach faces critical scalability limitations due to (1) increased annotation errors caused by 
inherent recognition complexity in low-resolution data, and (2) high costs and labor intensity (e.g., ~\$800 for annotating six hours of video footage~\cite{dataprice}).
Recent advances in synthetic data generation~\cite{long2024llms,patel2024datadreamer} suggest an alternative paradigm where LVLMs assist in automated caption generation. However, we have two observations: First, directly applying state-of-the-art captioning models like Tarsier~\cite{wang2024tarsier, yuan2025tarsier2} to raw low-resolution inputs yields semantically inconsistent descriptions, even with explicit text prompting (see comparative analysis in \S\ref{sec:exp:performance_on_aux}). Second, we identify that effectively leveraging class-aware guidance, e.g., pseudo labels, which strategically incorporate behavioral taxonomy context during inference, can significantly improve caption precision (by relatively 38.7\% in our measurements) through better extraction of concrete human behavior rather than vague sensor patterns.


Motivated by these observations, this paper aims to develop a system for HBU on low-resolution vision cameras. However, based on our deployment experience on a real-world low-resolution system, we have identified two critical challenges. 
The first challenge lies in accurately identifying individuals in long, low-resolution video frames at a low cost, namely \textit{temporal mismatch}. Existing methods on this issue either resort to learned ML models (e.g., YOLO~\cite{yolov8_ultralytics} and InFi~\cite{yuan2022infi}) that require massive computation GFLOPs (as measured in \S \ref{sec:exp:on-camera-filtering}), or apply on-device filtering approaches like Reducto~\cite{li2020reducto}, which are primarily designed for RGB cameras with less effectively in low-resolution vision sources. The second challenge is accurate human positioning that crops the human objective from low-resolution frames. This challenge arises from heterogeneous system configurations, such as variations in the view and position of low-resolution cameras.
This leads to a \textit{spatial mismatch}, where the model struggles to direct attention to the correct areas, compounded by the sparse information density of the low-resolution sensors. Unfortunately, we observe that these two mismatches significantly affect the overall system performance (as measured in \S \ref{sec:exp:effect_labeler}).

Therefore, to solve the temporal mismatch, we design a lightweight window-based sensitivity filtering mechanism (\S\ref{sec:method:stage1:step1}) that detects human presence via human-centric temporal features bypassing resource-intensive detectors while overcoming the limitations of RGB-centric filters. Next, to solve the spatial mismatch, we develop a dynamic action-capturing mechanism, which guides the model to focus on behavior-relevant regions (\S\ref{sec:method:stage1:step2}). Then, to obtain a well-trained labeler that generates class-aware pseudo labels, we design a contrastive labeler training mechanism to effectively leverage both labeled and unlabeled data (\S \ref{sec:method:stage1:step3}). In the following, to effectively guide Large Language Models (LLMs) in generating high-quality video-level captions using the pseudo labels, we propose spatial consistency and temporal coherence checks across frames to minimize potential annotation errors from the labeler (\S\ref{sec:method:stage2}).
Lastly, to ensure deployability, we further introduce a LoRA-based fine-tuning strategy that adapts LVLMs to low-resolution data with minimal parameter overhead (<3\%), balancing caption quality and computational efficiency (\S \ref{sec:method:stage3:lora}).

By integrating the above designs, we propose \workname, a novel system designed to enable low-resolution HBU on mobile devices with LVLMs.
\workname~introduces three key stages, namely \textit{Contrastive-Oriented Data Labeler},  \textit{Physical-Knowledge-Guided Captioner}, and \textit{LoRA-based efficient Fine-tuning}.
We evaluate \workname~on a region-scale testbed and three public datasets, showing it outperforms several state-of-the-art LVLM systems.  The main contributions are summarized as follows:
\begin{itemize}[leftmargin=*]
    \item We design \workname, a novel, labor-saving, three-stage on-device system designed to enable low-resolution HBU. The key insight of this work is that class-aware guidance can enhance caption precision, which significantly mitigates the limitations of current LVLM systems in understanding low-resolution data.
    \item To relieve the temporal and spatial mismatches inherent in low-resolution data for HBU, we propose a window-based sensitivity on-camera filtering mechanism along with fine-grained human action capturing. In addition, to enable LVLMs to better interpret low-resolution data and generate accurate captions, we introduce intra- and inter-distribution checking through spatial and temporal consistency.
    \item Extensive experiments show that \workname~achieves up to $87.02\%$ performance in Bert-Score (F1-score) in our real-world testbeds and three distinct low-resolution datasets with a relative improvement of $40.03\%$ on average compared to several state-of-the-art systems.
\end{itemize}

%% file: 3_application.tex
\section{Applications and  Motivations}
\label{sec:motivation}

In this section, we present the application scenarios and motivation of \workname. In particular, we explain why captions are better suited for representing HBU compared to label-based approaches. In the following, we discuss the limitations of existing LVLMs. 

\subsection{{Application Scenarios}}

\workname~can be applied to various scenarios involving low-resolution cameras, such as depth and thermal sensors, which are used to continuously and longitudinally monitor and understand user behavior~\cite{jiang2024artfl,yang2024drhouse}. For instance, in Alzheimer’s Disease monitoring applications~\cite{chen2019developing,ouyang2024admarker}, low-resolution cameras can provide persistent monitoring of an elderly individual’s daily activities, such as standing, walking, and engaging in social interactions. These activities serve as critical information for the early diagnosis of Alzheimer’s Disease and cognitive decline~\cite{ouyang2022cosmo}.  In addition, other health-related scenarios, such as Parkinson’s disease (PD)~\cite{sun2024digital} and rehabilitation training~\cite{touhiduzzaman2025wi}, also require reliable human activity understanding as a fundamental biomarker for advanced health monitoring.
Moreover, \workname~enables applications such as crowd monitoring~\cite{miao2022abnormal} and anomaly detection~\cite{zhang2021unsupervised} by using low-resolution sensors to track human presence and understand human activities.

\subsection{Limitation of Labels}
\label{sec:motivation:labels}
\begin{figure}
    \centering
    \includegraphics[width=.98\linewidth]{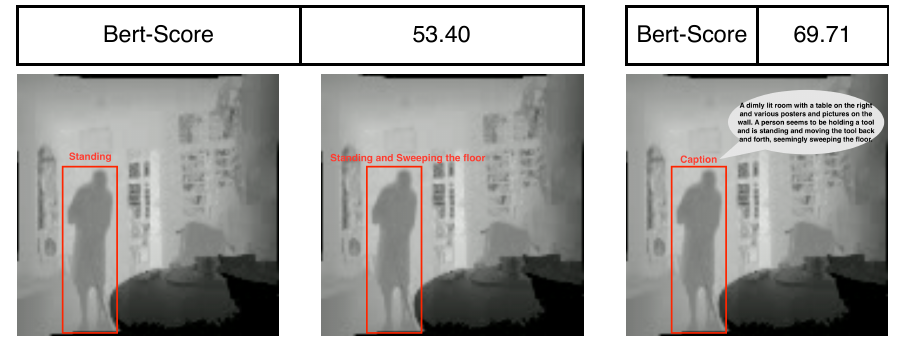}
    \vspace{-10pt}
    \caption{Limitation of labels. The left depth frame illustrates a single-label classification, while the middle depth frame represents a multi-label classification.  In contrast,  the right depth frame provides a caption with a detailed description of the scene.
    }
    \vspace{-10pt}
    \label{fig:motivation:label_limited}
\end{figure}

We present the limitations of describing human behaviors with labels, including single labels and multiple labels. 
As shown in Fig.~\ref{fig:motivation:label_limited}, it illustrates the limitations of single-label and multi-label classification compared to captioning. 
In particular, the left portion of the figure labeled ``Standing'' provides only a simplistic description, failing to capture concurrent actions, i.e., sweeping the floor. 
The middle portion, labeled ``Standing and Sweeping the Floor'', improves by identifying multiple actions but still lacks details about the surrounding environment, such as the presence of furniture or objects. 
In contrast, the right portion uses captioning to deliver a comprehensive description: ``\textit{A dimly lit room with a table on the right and various posters and pictures on the wall. 
A person seems to be holding a tool and is standing and moving the tool back and forth, seemingly sweeping the floor.}'' This caption recognizes the actions, offering a richer understanding of the scene. 
In addition, we conduct preliminary experiments comparing the effectiveness of captions versus labels in adapting general LVLMs to the specific HBU domain.
As shown in Fig.~\ref{fig:motivation:label_limited}, fine-tuning an LVLM, i.e., Qwen-VL \cite{Qwen2.5-VL}, using captions achieves a Bert-Score (F1-Score) of $69.71\%$, whereas fine-tuning with labels achieves only $53.40\%$.
This indicates that captions are more effective in utilizing the capabilities of LVLMs in HBU by providing a more holistic and detailed representation, which is crucial for many real-world applications requiring both actions and context, such as activity monitoring and eldercare.



\subsection{Limitation of Existing LVLMs} 
\subsubsection{Task gap in HBU}
\label{sec:motivation:taskgap}
Current LVLMs are predominantly designed for general visual question-answering tasks on RGB-text data~\cite{bai2023qwen,Deepseek-VL2,yang2024viassist}, which often lack specialization for specific domains such as HBU. Towards this challenge, we conduct an empirical study on the UTD dataset~\cite{chen2015utd} using RGB videos for human behavior understanding. We evaluate four baselines, i.e., Qwen2.5-VL-7B~\cite{Qwen2.5-VL} and InternVL2-8B~\cite{chen2024internvl}, OneLLM-7B~\cite{han2023onellm} and LanguageBind-7B~\cite{zhu2023languagebind}. We use accuracy as the evaluation metric, measuring the models' ability to correctly select the appropriate label from a predefined list of actions.  For comparison, we establish an upper-bound performance using task-specific models trained on the full dataset, achieving an accuracy of $92.23\%$ in RGB video~\cite{xu2023mesen}.  As shown in Fig.~\ref{fig:motivation:limit_LLMs_rgb}, the performance of four models falls significantly below the upper bound, with the highest accuracy being $25.92\%$. This underscores the substantial task gap that exists in HBU.

\subsubsection{Modality gap in low-resolution vision data}
\label{sec:motivation:modalgap}
\begin{figure}[t]
    \centering
    \begin{subfigure}[b]{0.23\textwidth} 
        \centering
        \includegraphics[width=.92\textwidth]
        {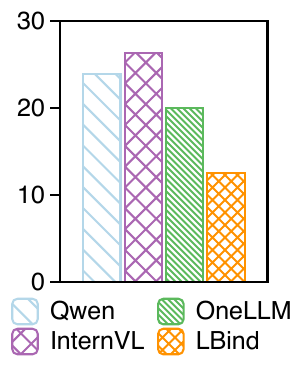} 
        \vspace{-8pt}
        \caption{RGB Performance}
        \label{fig:motivation:limit_LLMs_rgb}
    \end{subfigure}
    \hfill
    \begin{subfigure}[b]{0.23\textwidth} 
        \centering
        \includegraphics[width=.92\textwidth]{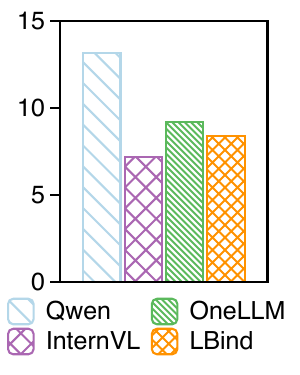} 
        \vspace{-8pt}
        \caption{Depth Performance}
        \label{fig:motivation:limit_LLMs_depth}
    \end{subfigure}
    \vspace{-8pt}
    \caption[]{Accuracy performance of the existing LVLMs. Existing single-modality LVLMs, such as Qwen2.5-VL-7B (Qwen) and InternVL2-8B (InternVL), as well as multi-modality LVLMs, such as OneLLM-7B (OneLLM) and LanguageBind-7B (LBind), demonstrate limited performance in understanding human behavior.}
    \label{fig:motivation:limit_LLMs}
    \vspace{-10pt}
\end{figure}

The second critical challenge is the modality gap when dealing with low-resolution vision data. Most LVLMs such as Qwen2.5-VL-7B and InternVL2-8B, are primarily designed and trained for RGB data, which limits their ability to effectively interpret depth information. Some LVLMs designed to handle multiple modalities, e.g., OneLLM-7B and LanguageBind-7B, can support captioning with depth data.
Their performance on depth-based tasks, however, remains suboptimal compared to task-specific models. Under the same settings in \S\ref{sec:motivation:taskgap}, we train a task-specific model on the full dataset, observing an accuracy of $82.30\%$ for depth video classification. 
Fig.~\ref{fig:motivation:limit_LLMs_depth} clearly illustrates the detrimental impact of such a modality gap with the highest accuracy being $14.81\%$. 


\subsection{Summary}
To address the challenges of understanding low-resolution data in HBU, our motivation study underscores that directly leveraging existing LVLMs can not perform well. However, the use of ``⟨low-resolution data, caption⟩'' pairs has been shown to significantly mitigate these limitations. However, obtaining a sufficient quantity of such pairs remains a significant challenge due to the labor-intensive nature of annotation. Furthermore, several challenges, such as temporal and spatial mismatches, need to be addressed due to the inherent discrepancies between high-resolution and low-resolution data.

%% file: 4_overview.tex
\section{System Overview}
\label{sec:system_overview}
In this section, we first present the problem formulation.
Then, we introduce the system architecture of \workname.
\subsection{Problem  Formulation}

We consider a captioning task for HBU with low-resolution videos, collected by low-resolution sensors such as depth, thermal, and infrared.
A low-resolution video can be represented as a sequence of frames, defined by $\mathbf{V}=\{I^{V}_1, \ldots, I^{V}_{n_{V}}\}$, where $n_{V}$ denotes the total number of frames. 
The frames are partitioned into two distinct subsets. One is the set of frames without presence of people, denoted as $\mathbf{V}_e=\{I^{V}_{i}\mid i \in \mathcal{I}_e\}$; the other is the set of frames containing human behavior-relevant content, denoted as $\mathbf{V}_h=\{I^{V}_{j}\mid j \in \mathcal{I}_h\}$.
$\mathcal{I}_e$ and $\mathcal{I}_h$ represent the corresponding sets of frame indices satisfying $\mathcal{I}_e \cup \mathcal{I}_h = \{1,\ldots,n_{V}\}$ and $\mathcal{I}_e \cap \mathcal{I}_h = \emptyset$. 
For HBU tasks, the subset $\mathbf{V}_h$ is the primary focus.
To facilitate effective analysis, $\mathbf{V}_h$ can be summarized using video captions, denoted as $\mathbf{C}_h$, which comprehensively describes the human activities present in $\mathbf{V}_h$.
In this work, limited labeled videos $\mathbf{V}^\text{l}$ and abundant unlabeled videos $\mathbf{V}^\text{u}$ are available, all collected by various low-resolution sensors. \workname~aims to enhance LVLMs’ capabilities in HBU with the video-caption pairs ⟨$\mathbf{V}_h$, $\mathbf{C}_h$⟩ obtained from the collected low-resolution videos $\mathbf{V}$, thus further enabling effective on-device HBU applications.

\subsection{System Architecture}
\begin{figure*}
    \centering
    \includegraphics[width=.98\linewidth]{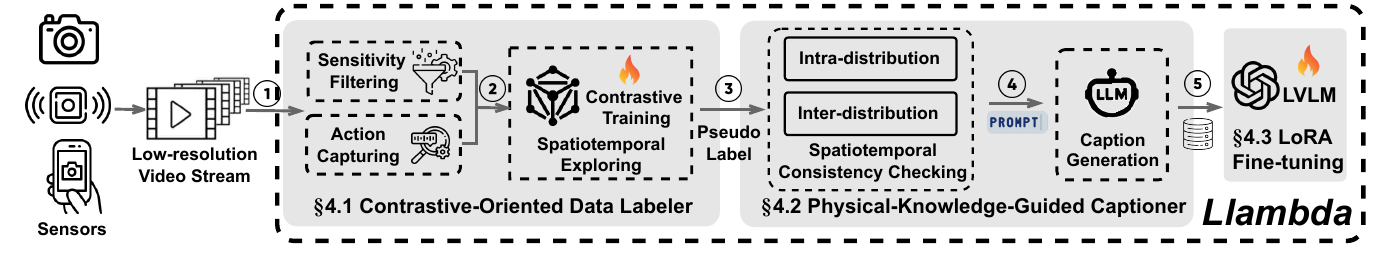}
    \vspace{-12pt}
    \caption{Overview of \workname. \workname~ has three stages including Contrastive-Oriented Data Labeler, Physical-Knowledge-Guided Captioner, and LoRA-based Efficient Fine-tuning.}
    \label{fig:enter-label}
\end{figure*}

\workname~is designed to support HBU tasks based on low-resolution data with limited labels. 
Fig.~\ref{fig:enter-label} demonstrates the pipeline of \workname, which mainly contains three stages, including \textit{Contrastive-Oriented Data Labeler}, \textit{Physical-Knowledge-Guided Captioner}, and \textit{LoRA-Based Efficient Fine-Tuning}.

\textbf{(1)} In \textit{Contrastive-Oriented Data Labeler},  \workname~first removes irrelevant or noisy data from the input low-resolution video streams through a window-based sensitivity filtering method followed by a dynamic action capturing.
\textbf{(2)} Then,  we use the selected data with contrastive learning approaches to enhance the capability of our labeler, which can subsequently be used to annotate unlabeled data with class-aware pseudo labels, i.e., the logits of classification. Note that, in practice, the training of the labeler can be performed in either a local mode or a federated mode, depending on the amount of data or the user's configuration. 


\textbf{(3)} In \textit{Physical-Knowledge-Guided Captioner}, once we obtain pseudo labels for each frame of the selected unlabeled data via the labeler, \workname~aims to utilize them to generate action-specific video captions with LLMs. However, these frame-level pseudo labels usually constitute high-dimensional time-series data at the video level.
LLMs struggle to process them effectively due to their limited ability in handling large-scale numerical data, which often results in inaccurate captions.
To address this issue, we first propose spatiotemporal consistency checking, which evaluates both the intra-distribution and inter-distribution of the data to ensure label reliability and accuracy. \textbf{(4)} After the checking, the LLMs receive the prompt along with top-K information of pseudo labels and generate descriptive, video-level captions. These generated captions enable \workname~to provide a high-level understanding of the observed behaviors from the input low-resolution video stream.
 
\textbf{(5)} In \textit{LoRA-based Efficient Fine-tuning}, \workname~leverages the raw low-resolution data and generated captions to fine-tune pre-trained LVLMs using Low-Rank Adaptation (LoRA)\cite{hulora}, offering minimal computational overhead and being edge-friendly. By integrating these three designed stages, \workname~enables accurate and efficient human behavior understanding for low-resolution data in environments with constrained resources.


%% file: 5_design.tex
\section{System Design}
\label{sec:method}
We propose a novel three-stage framework that leverages a limited amount of labeled data, a large volume of unlabeled data, and the knowledge of LLMs to support HBU tasks with low-resolution modalities.

\subsection{Contrastive-Oriented Data Labeler}
\label{sec:method:stage1}
In the first stage of \workname, our goal is to train a labeler to generate pseudo labels for the unlabeled low-resolution sensor data. The key idea behind our design is to first select high-quality data and then combine the knowledge from the limited labeled data with the large volume of unlabeled data in a contrastive manner. 
\subsubsection{Window-based sensitivity on-camera filtering}
\label{sec:method:stage1:step1}

\label{sec:method:ocf}
\begin{figure}
    \centering
    \includegraphics[width=\linewidth]{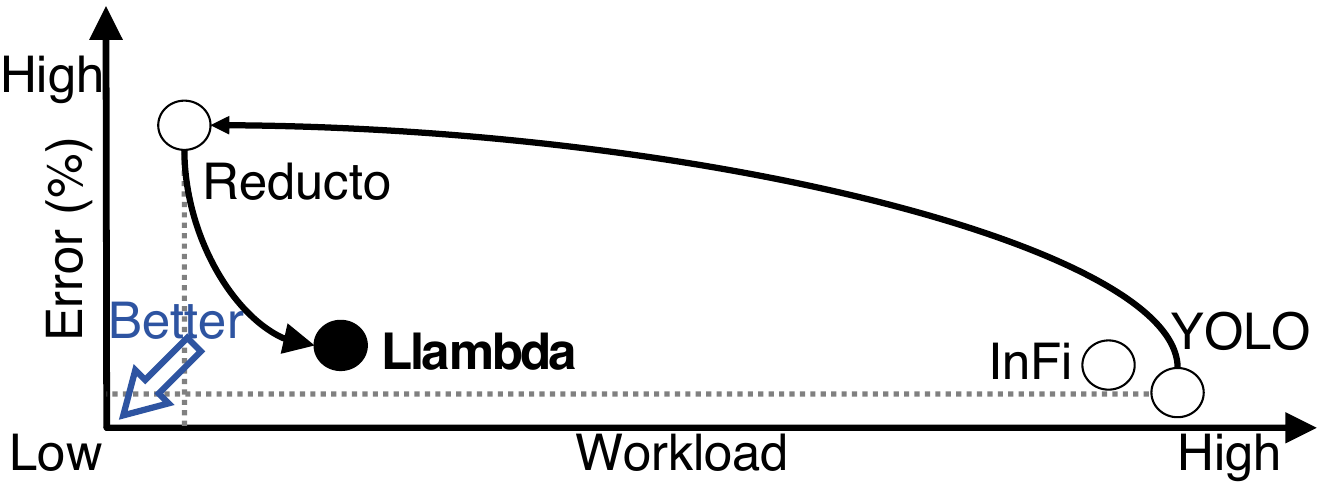}
    \vspace{-15pt}
    \caption{Comparison of filtering approaches including Reducto~\cite{li2020reducto}, InFi~\cite{yuan2022infi}, YOLO~\cite{yolov8_ultralytics} in terms of error rate and computational workload (GFLOPs).}
    \vspace{-10pt}
    \label{fig:method:motivation1}
\end{figure}
We aim to filter useful unlabeled data from low-resolution videos, as fixed monitoring devices in HBU often miss individuals, making processing all frames inefficient.
For instance, a device placed in a living room may mostly record frames without the presence of people at night, since people typically sleep in the bedroom.  A common approach to address this issue is to use a detection model, such as YOLO~\cite{redmon2017yolo9000}, to identify the presence of people in each frame and extract relevant video clips.  However, this method involves significant computational overhead, as the detection model must run continuously to process all frames. Another approach is to train a binary classifier~\cite{yuan2022infi} to determine whether a person is within the field of view.
Although this method reduces inference costs compared to detection models, it requires considerable manual effort to label data in order to train a high-accuracy classifier. 

Therefore, there is a strong need for a more cost-effective solution to efficiently select video clips containing behavior-relevant content without relying on resource-intensive and continuous inference. Inspired by Reducto~\cite{li2020reducto}, we propose a window-based sensitivity on-camera filtering method to leverage simple patterns, e.g., cheap vision features, for coarse-grained frame filtering.  As illustrated in Fig.~\ref{fig:method:motivation1}, we hope that our approach aims to operate in the lower-left region of the error-workload space, achieving high accuracy (close to YOLO) in detecting people while maintaining low computational overhead (close to Reducto).  Therefore, we propose a sliding window mechanism with sensitivity adjustments to efficiently identify those frames containing people.

Let $I_{t}^{V}$ represents the frame at time $t$ from the video $\mathbf{V}$.
A window containing a sequence of consecutive frames is defined as $\mathbf{W}_t = \{{I}_{t}^{V}, {I}_{t+1}^{V}, \ldots, {I}_{t+w-1}^{V}\}$, where $w$ is the window size.
For each window $\mathbf{W}_t$, we extract simple features, such as pixel differences, to assess the presence of people. Let $\mathbf{D}_t$ represent the set of pixel differences computed from $\mathbf{W}_t$, defined as: $\mathbf{D}_t = \{d_i = F({I}_{i+1}^{V}, {I}_{i}^{V}) \ | \ i \in [t, t+w-2]\}$,
where $F(I_{i+1}, I_{i})$ computes the pixel-wise difference between two consecutive frames. Next, we define a binary decision function, denoted as $\mathcal{C}$, to determine whether a frame should be retained. For each pixel difference $d_i \in \mathbf{D}_t$, we compute a binary score $S_i$ as follows:
\begin{align}
S_i =
\begin{cases}
1, & d_i > \sigma \cdot d_{\text{max}}, \\
0, & \text{otherwise},
\end{cases}    
\end{align}
where $\sigma$ is a threshold for detecting significant changes, and $d_{\text{max}}$ is the maximum pixel difference observed in $\mathbf{D}_t$. 

Finally, if the decision function $\mathcal{C}(\mathbf{S})$, which sums the elements of $\mathbf{S} = \{S_i\}$, satisfies the condition $\mathcal{C}(\mathbf{S}) \geq N$, current frame ${I}_{i}^{V}$ in the window $\mathbf{W}_t$ would not be retained. Here, $N$ is a hyperparameter that defines the minimum number of significant differences required to keep a frame. In our experiments, we set $N = 2$.

The core insight behind the design is that we aim to detect two specific situations in HBU, i.e., noise and the presence of a person entering the scene. In the first situation, noise may arise due to accidental environmental factors or device errors, which are common in low-resolution sensors. Such noise typically manifests as sporadic large pixel differences between adjacent frames, while the pixel differences in other frames remain relatively small and inconsistent. In contrast, when a person enters the scene, the pixel differences between consecutive frames exhibit a gradual, sequential decrease over time, without abrupt peaks. This distinct pattern allows us to effectively differentiate human movement from noise. 

\subsubsection{Fine-grained human action capturing}
\label{sec:method:stage1:step2}
\begin{figure}
    \centering
    \includegraphics[width=0.95\linewidth]{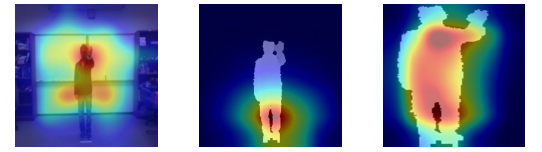}
    \vspace{-5pt}
    \caption{Motivation of dynamic action capturing. We highlight the importance of focusing attention in action recognition using RGB and depth data. By applying a cropping mechanism, the model can better concentrate on human-relevant regions, reducing background noise and improving recognition performance.}
    \vspace{-10pt}
    \label{fig:motivation_recovery}
\end{figure}

Compared to high-resolution data, low-resolution data, such as depth, thermal, or infrared data, carries significantly less information, making it more challenging to extract meaningful features. This limitation indicates the need for additional labeled data to enable the model to converge effectively. To mitigate this issue, it is crucial to guide the model's attention to the most relevant regions about people of the input data. 
In particular, we conduct an experiment on the UTD-D dataset~\cite{chen2015utd} to demonstrate the impact of low-resolution data on pretrained models' attention to regions of interest. 
Under the same conditions and training data, RGB-trained models focus better on correct regions (Fig.\ref{fig:motivation_recovery}, first), while depth-trained models struggle (Fig.\ref{fig:motivation_recovery}, second).
However, by cropping the relevant regions about the target subject in the depth data using prior information, we can effectively guide the model's attention to the appropriate areas.
This approach is particularly effective when labeled data is limited, as it helps improve the model's performance by mitigating the challenges associated with low-resolution inputs.

Therefore, to effectively obtain and crop the relevant regions about people, we incorporate YOLO-V8~\cite{yolov8_ultralytics} to detect and localize individuals in the selected video clips. 
In our approach, the detector is applied only to frames that have been pre-filtered by the window-based sensitivity method according to the \S\ref{sec:method:ocf}, thereby reducing computational overhead while maintaining high detection accuracy. In particular, for the $t$-th frame ${I}_{t}^{V}$ in the video $\mathbf{V}$, we have $\mathbf{y} = f({I}_{t}^{V}; \theta)$, where $\theta$ represent the parameters of the detection model. $\mathbf{y}$ is a set of predicted bounding boxes and associated confidence scores with $\mathbf{b} = \left(x, y, w, h, p\right)$. Note that  $(x, y)$ is the coordinates of the box center, $(w, h)$ are the width and height, and $p$ denotes the detection probability for the human class. 
Next, we further analyze the spatial positions and temporal coherence of the bounding boxes across consecutive frames. This allows us to capture both the appearance and motion patterns of human subjects which can be formulated as
\begin{align}
\Delta \mathbf{b}_t = \lVert \mathbf{b}_{t} - \mathbf{b}_{t+1} \rVert_{2} < \epsilon, \notag
\end{align}
where $\epsilon$ specifies the maximum allowable displacement between bounding boxes in adjacent frames.
This ensures that only consistent and meaningful detections are utilized for downstream analysis, minimizing the impact of missed detections resulting from low-resolution data. Note that we can fine-tune the detection model if the input resolution is too low for the model to effectively detect objects. 
In practice, we used only $1\%$ of labeled depth data to fine-tune extremely low-resolution datasets, achieving satisfactory performance (see comparative analysis in \S~\ref{sec:exp:vis_hac}).

\subsubsection{Explicit spatiotemporal exploring via contrastive learning}
\label{sec:method:stage1:step3}


After selecting unlabeled data from low-resolution video frames, we train a labeler with limited labeled and unlabeled data in contrastive learning. A direct solution is to use the video-label pair to train a video labeler. However, the video-wise labeler lacks detailed spatiotemporal information, limiting its effectiveness in providing such details to the captioning model (\S \ref{sec:method:captioner}). Thus, we split video clips into individual frames to actively explore spatiotemporal information, thereby enhancing the training process. Specifically, we adopt the NT-Xent loss~\cite{jiang2022pgada} into the overall training framework, which is defined as follows:
\begin{align}
\mathcal{L}_{\text{C}} = - \frac{1}{N} \sum_{i=1}^N \log 
\frac{\exp\left(\text{sim}(\mathbf{z}_i, \mathbf{z}_i^+) / \tau \right)}{
 \sum_{j \neq i} w_{ij} \exp\left(\text{sim}(\mathbf{z}_i, \mathbf{z}_j) / \tau \right)},
\end{align}
where $\text{sim}(\cdot, \cdot)$ denotes the cosine similarity between two feature vector. $\mathbf{z}_i$ and $\mathbf{z}_i^+$ are the representations of a positive pair (e.g., augmented views of the same sample). $\mathbf{z}_j$ represents a negative sample. $\tau$ is the temperature scaling parameter. $w_{ij}$ is a semantic weight that modulates the contribution of the negative sample $\mathbf{z}_j$, computed based on the semantic similarity between $\mathbf{z}_i$ and $\mathbf{z}_j$. In addition, to enhance the classification capability of the model, we adopt the cross-entropy loss $\mathcal{L}_{CE}$. Thus, the total training loss combines the semantic-aware contrastive loss for self-supervised learning and the cross-entropy loss for supervised learning, as shown below:
\begin{align}
\mathcal{L} = \lambda \mathcal{L}_{\text{C}} + (1 - \lambda) \mathcal{L}_{\text{CE}},
\end{align}
where $\lambda$ is a hyperparameter to balance the two losses.

However, when deploying our system on edge devices, the distributed nature of the data arises from the limitations of labeled data and privacy constraints. As a result, \workname~can also support federated learning to address scenarios where data cannot be centralized, as formulated below.
\begin{align}
\min_w \sum_{i=1}^N \frac{|D_i|}{\sum_{j=1}^N |D_j|} \left( \frac{1}{|D_i|} \sum_{(x, y) \in D_i} \mathcal{L}(w, x, y) \right),
\end{align}
where, $N$ denotes the total number of participating devices, \( D_i \) denotes the local dataset on client \( i \). $ | \cdot | $ denotes the instance number of the local datasets. $x$ and $y$ denote the data and the label. $\mathcal{L}$ denotes the training loss and $w$ is the model weight. For simplicity, we adopt the vanilla FedAvg~\cite{mcmahan2017communication} algorithm in our aggregation process while \workname~can allow for the integration of more advanced federated learning aggregation techniques, such as FedProx~\cite{yuan2022convergence} and ArtFL~\cite{jiang2024artfl}.



\subsection{Physical-Knowledge-Guided Captioner}
\label{sec:method:stage2}



\begin{figure}
    \centering
\includegraphics[width=\linewidth]{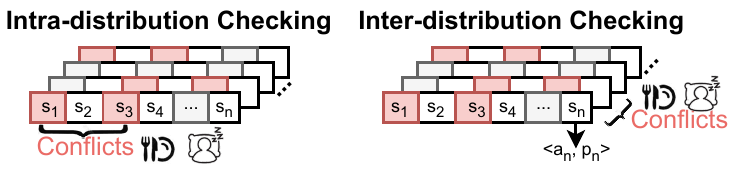}
    \vspace{-20pt}
    \caption{Illustration of intra- and inter-distribution checkings.}
    \vspace{-10pt}
    \label{fig:checking}
\end{figure}

%
\label{sec:method:captioner}
Once obtaining a well-trained labeler, we can use it to annotate unlabeled videos
with class-aware pseudo labels. However, understanding high-dimensional time-series data composed of frame-level pseudo labels remains challenging for LLMs due to potential errors and the inherent complexity of the data (as analyzed in \S\ref{sec:exp:top-k}). In this section, we introduce a physical-knowledge-guided captioner that integrates both intra- and inter-distribution checking to address the challenges of spatial and temporal consistency in the labeler's predictions, respectively. As illustrated in Fig.~\ref{fig:checking}, the intra-distribution checking focuses on identifying action conflicts within each pseudo label, while the inter-distribution checking addresses conflicts across the sequence of pseudo labels.



\subsubsection{Intra-distribution checking via spatial consistency}
\label{Intra-distribution Checking via Spatial Consistency}

Human activities in the real world often involve multiple actions occurring simultaneously.
Directly sending the sequence of predictions into an LLM for reasoning will result in inferior captioning performance due to spatial conflicts in the predictions generated by the labeler in \S\ref{sec:method:stage1}.
One straightforward approach to addressing the challenge of spatial consistency is to input the distribution of the labeler's predictions (i.e., pseudo labels) into the LLM for reasoning, instead of using hard labels.
However, the large quantity of numerical data is difficult for LLMs to comprehend, as various studies have demonstrated the inferior capabilities of LLMs in solving mathematical problems~\cite{yuan2023scaling}.

To address these challenges, we developed an uncertainty-checking-based filtering approach for inter-distribution validation.
Specifically, we format the predictions of the labeler for each frame by time, denoted as $[\bm{s1}, \bm{s2}, ..., \bm{s_n}]$.
The state $\bm{s_t}$ at each moment consists of its top-k categories and corresponding probabilities, i.e., $s_t = \{\bm{a_1}, \bm{p_1}, \dots, \bm{a_k}, \bm{p_k}\}$, where $\bm{a_i}$ and $\bm{p_i}$ are the predicted activities and probabilities.
We then pass the top-k behavior predictions and their corresponding probabilities for each frame to the LLM for reasoning. Here, $k$ is a hyperparameter whose impact and optimal value will be discussed in \S\ref{sec:exp:top-k}.
This approach enables the LLM to understand the events occurring throughout the entire video at a high level, resulting in captions that contain more comprehensive information.

\subsubsection{Inter-distribution checking via temporal consistency}
\label{sec:method:stage2:step2}
In addition to spatial conflicts in the activity sequences, there are also temporal conflicts in the activity prediction sequence annotated by the labeler. For instance, due to the limited capabilities of the labeler, which performs frame-wise activity recognition, there may be a ``running'' activity spanning 10 consecutive frames with a ``sleeping'' activity interspersed between them, which is counterintuitive and logically inconsistent.
To address this issue, we utilize a rule-based inter-distribution checking approach. Specifically, we set up a template to filter the inconsistency between multiple consecutive frames. 
Finally, we input the action sequence that has filtered out the temporal and spatial conflicts into the LLMs for inference to obtain the captions of the entire video. 

\subsubsection{Prompt-instructed caption generation}
\label{sec:method:stage2:step3}
This subsection introduces the details of the prompt used in the captioner. 
Specifically, the prompt consists of two parts: the system prompt and the runtime prompt. The system prompt remains static and includes task instructions, spatial consistency, and temporal consistency checks. The runtime prompt contains predictions from the labeler (see \S\ref{Intra-distribution Checking via Spatial Consistency}). After checking for spatial and temporal consistency, the LLM generates high-quality captions for the input low-resolution video data. The paired video data and captions are then used for LVLMs fine-tuning, which can be seen in \S\ref{sec:method:stage3:lora}.

\subsection{LoRA-based Efficient Fine-tuning}
\label{sec:method:stage3:lora}
To enable efficient on-device training for large models, especially in scenarios requiring real-time personalization or adaptation to local data, we utilize the Low-Rank Adaptation (LoRA) technique~\cite{hulora}. LoRA provides a practical solution to fine-tune LLMs for mobile and edge computing applications. The primary objective is to adapt the model to specific tasks while preserving its general capabilities, all while minimizing computational and memory overhead. 
For a given layer in the neural network, let $\mathbf{W} \in \mathbb{R}^{d \times d}$ be the weight matrix. Instead of directly updating $\mathbf{W}$ during fine-tuning, we introduce two auxiliary matrices, $\mathbf{A} \in \mathbb{R}^{d \times r}, \quad \mathbf{B} \in \mathbb{R}^{r \times d}$, where $r \ll d$ is the rank of the decomposition. The effective weight matrix during fine-tuning becomes:
\begin{align*}
    \tilde{\mathbf{W}} = \mathbf{W} + \mathbf{A}\mathbf{B}
\end{align*}
where $\mathbf{A}$ and $\mathbf{B}$ are the only parameters that need to be updated during fine-tuning, significantly reducing the number of trainable parameters compared to standard fine-tuning.


%% file: 6_exp.tex
\section{Region-scale real-world testbed Evaluations}
\label{sec:real-world-testbed}

\subsection{Region-scale Real-world Testbed } 

\subsubsection{Description of real-world testbed}
\begin{figure}[t]
    \centering
    \includegraphics[width=.98\linewidth]{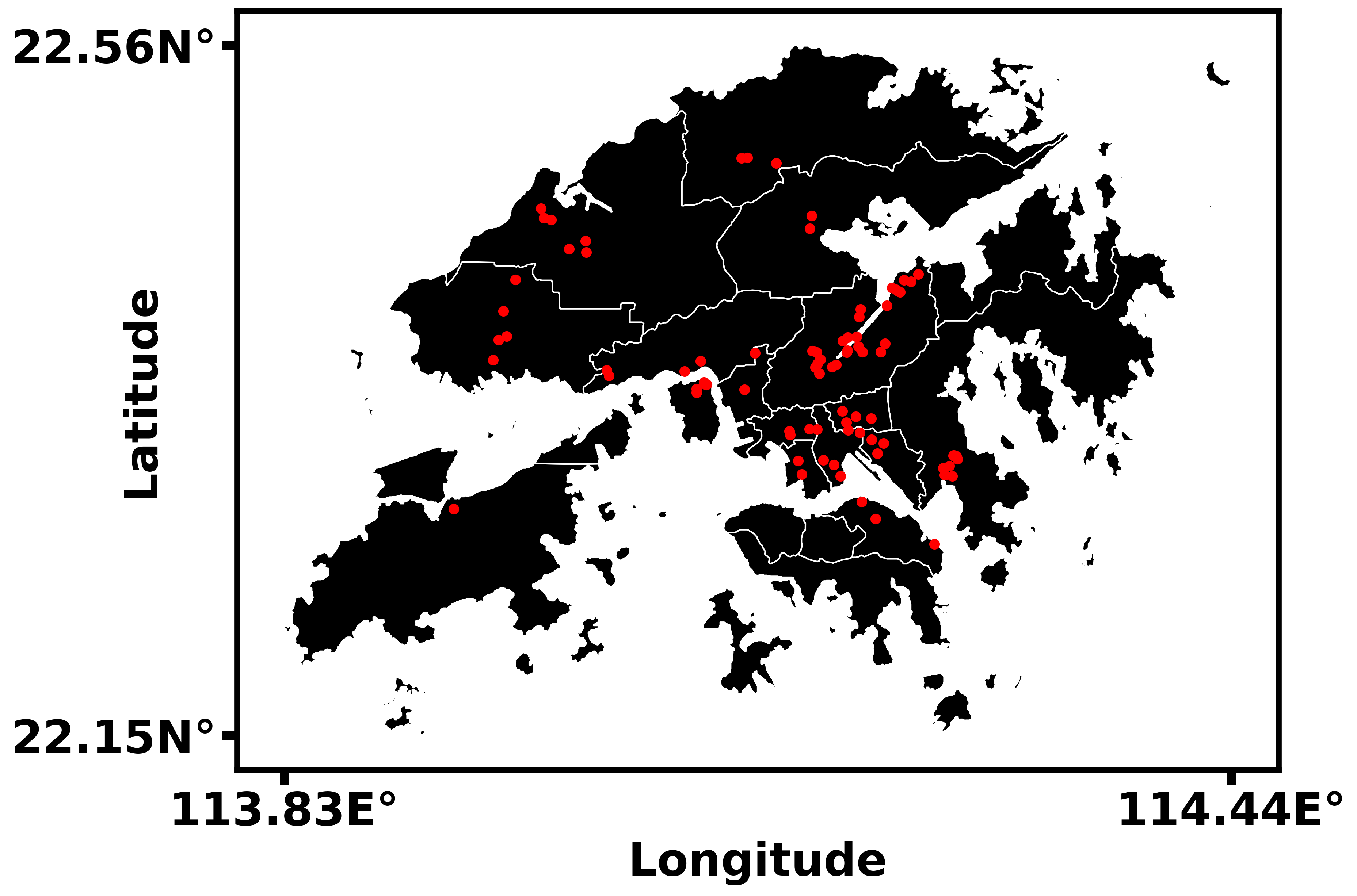}
    \vspace{-5pt}
    \caption{User distribution of our testbed. The red dot denotes the approximate position of users.}
    \label{fig:testbed:layout}
\end{figure}

We evaluate our system on a region-scale real-world testbed (RW-D) which is designed to analyze multiple digital biomarkers (e.g., activities of daily living, behavioral and psychological symptoms of dementia, and motor functions) to support Alzheimer's disease research in the homes of elderly individuals. In this work, we use a depth camera and a small amount of labeled data to capture 16 activities: ``Sitting'', ``Other actions'', ``Standing'', ``Walking'', ``Eating/Medication'', ``Grooming/Hair styling'', ``Exercising'', ``Handling objects'', ``Interacting/Socializing'', ``Sleeping/Lying down'', ``Transitioning (Sit/Stand)'', ``Using mobile phone'', ``Drinking'', "Cleaning'', ``Dressing/Undressing'', ``Rubbing/Washing hands''. As illustrated in Fig.~\ref{fig:testbed:layout}, the testbed consists of 97 sensor nodes deployed across participants' homes within an area of 1,115 $km^{2}$, with an average of one node per 11.49 $km^{2}$.  In addition, a central server is located in our lab. A sample of the collected data is presented in Fig.~\ref{fig:hardware:1}.

\subsubsection{Hardware setup}
\begin{figure}[t]
    \centering
    \begin{subfigure}[b]{0.23\textwidth} 
        \centering
        \includegraphics[width=\textwidth]{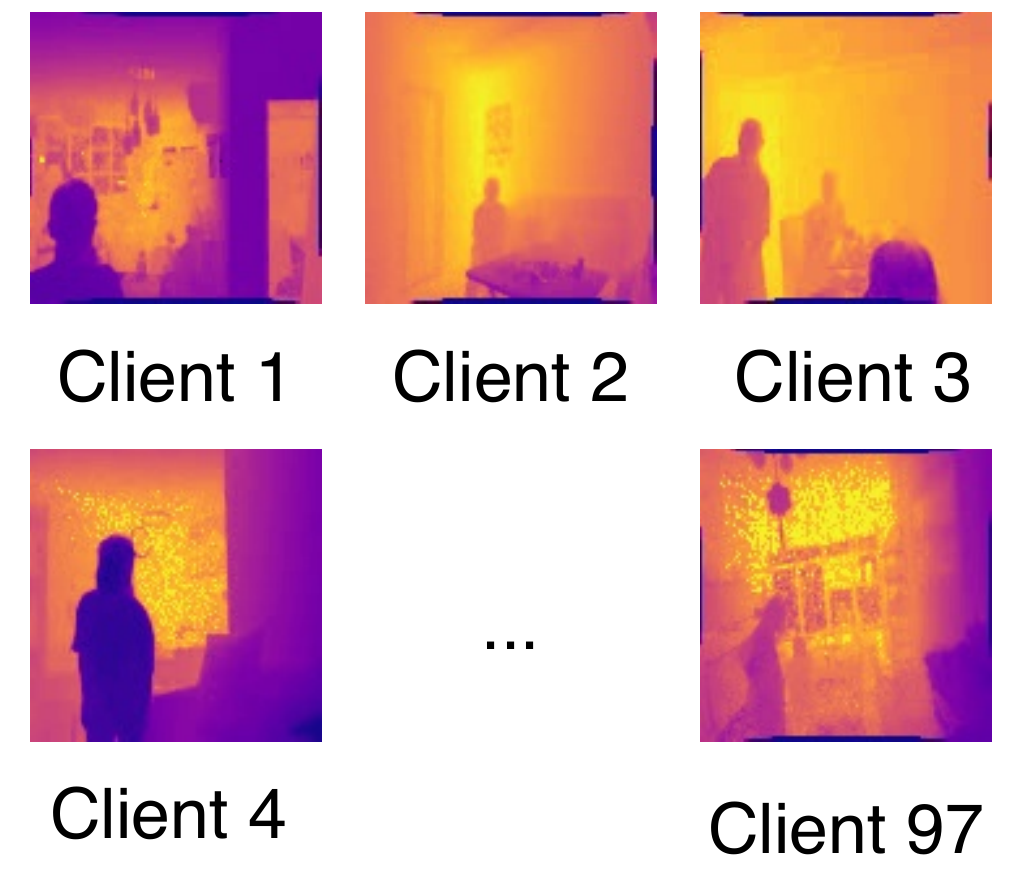} 
        \vspace{-5pt}
        \caption{Data Illustation}
        \label{fig:hardware:1}
    \end{subfigure}
    \hfill
    \begin{subfigure}[b]{0.23\textwidth} 
        \centering
        \includegraphics[width=\textwidth]{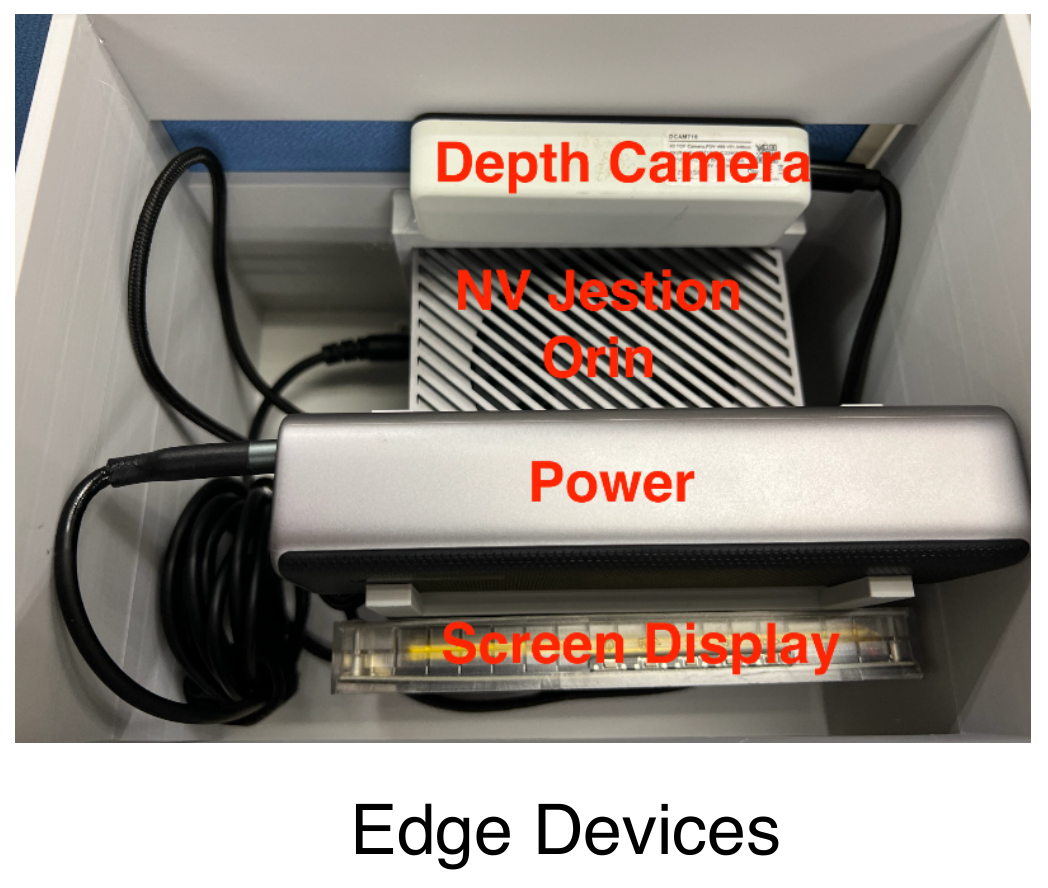} 
        \vspace{-5pt}
        \caption{Hardware Illustration}
        \label{fig:hardware:2}
    \end{subfigure}
    \vspace{-5pt}
    \caption{ Data example collected from our testbed and the server consists of a depth camera, Jetson Orin, power and screen display.}
    \label{fig:hardware}
\end{figure}

\begin{figure}[t]
    \centering
    \includegraphics[width=\linewidth]{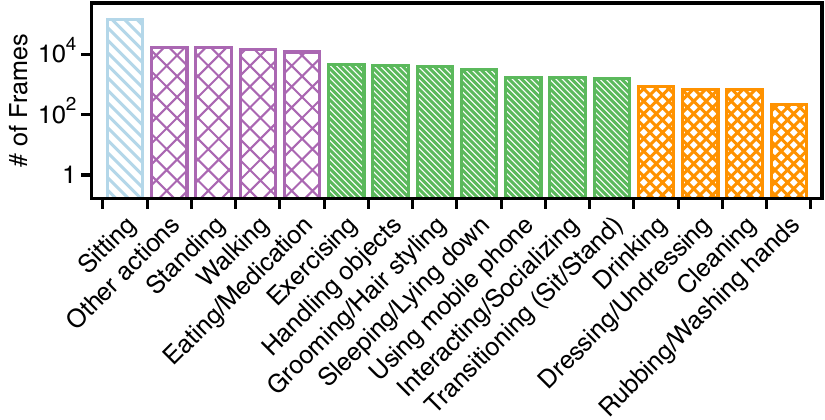}
    \vspace{-10pt}
    \caption{The distribution of labeled data of our testbed.}
    \label{fig:testbed:label_count}
\end{figure}

We set an NVIDIA Jetson Orin as a server in our lab, equipped with a depth camera. Additionally, we provide a power bank and a display screen to serve as a mobile power source, enabling movements, as shown in Fig.~\ref{fig:hardware:2}). Note that the server is connected to clients via Ethernet or Wi-Fi. Each sensor node is equipped with a hardware system that includes a depth camera, an NVIDIA Jetson Xavier NX single-board edge computer with a 1TB external NVMe SSD, and a 4G cellular interface for communication with the lab server. These nodes can collect and store depth data, locally train deep learning models, and communicate with the server as part of a federated learning framework. 
The NVIDIA Jetson Xavier NX edge devices (featuring a 6-core ARM CPU, 512 CUDA cores, and 16GB of memory) run on Ubuntu 18.04, while the lab server (equipped with a 12-core ARM CPU, 2048 CUDA cores, and 64GB of memory) also operates on Ubuntu 18.04. \workname~is deployed across these nodes and runs continuously for four weeks. As shown in Fig.\ref{fig:testbed:label_count}, only 1\% of the collected data is labeled, illustrating the distribution of the labeled dataset.

\subsection{Configurations of real-world testbed}
We present the baselines, evaluation metrics, and configuration details of our region-scale testbed.

\label{sec:testbed:configurations}
\subsubsection{Baselines} We evaluate \workname~on five state-of-the-art LVLMs systems including Qwen2.5-VL~\cite{Qwen2.5-VL}, Imagebind-LLM~\cite{han2023imagebind}, OneLLM~\cite{han2024onellm}, Language-bind~\cite{zhu2023languagebind}, and Tarsier~\cite{wang2024tarsier}. Specifically, Qwen2.5-VL is a state-of-the-art vision-language system designed to enhance the understanding and interaction between text and visual data. Imagebind-LLM is a multimodal system that bridges the gap between images and language using advanced binding mechanisms. OneLLM provides a unified framework for tackling various multimodal tasks, such as vision-language understanding, question answering, and image captioning. Language-bind focuses on improving the integration of language and vision models by establishing strong semantic connections between text and images. Lastly, Tarsier is an efficient captioning system that delivers strong performance on RGB data. It is important to note that Imagebind-LLM, OneLLM, and Language-bind are strong baselines since they can support depth modality. 

\subsubsection{Evaluation metrics and configurations details}  we use Bert-Score~\cite{zhang2019bertscore} as our metric for evaluating text generation quality that leverages the bidirectional context-aware embeddings from the BERT model. It calculates similarity scores between individual words of two texts, allowing for flexible alignment even when sentence structures differ. We use $F1$-score for our final results which consider the \textit{Precision} and \textit{Recall} simultaneously. For the ground truth, we recruited 3 volunteers to annotate captions for the data in our testbed. To ensure the quality of the captions, we provided the volunteers with detailed action instructions. The detail implemtation could be found in \S\ref{sec:exp:implement_details}.

\subsection{Results on real-world testbed}
\label{sec:exp:results-on-RW-D}
\subsubsection{Overall performance}
\begin{figure}[t]
    \centering
    \includegraphics[width=\linewidth]{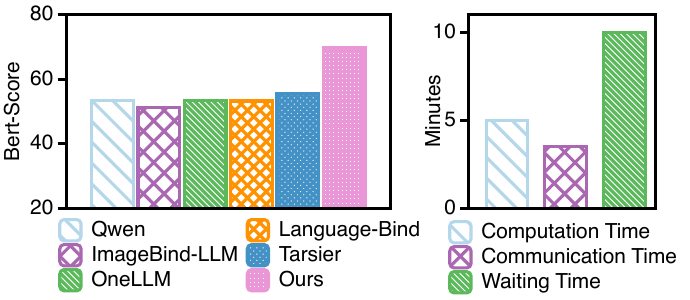} 
    \vspace{-10pt}
    \caption{(\textit{Left}): Performance (Bert-Score) comparison on RW-D testbed. \workname~consistently outperforms the state-of-the-art baselines. (\textit{Right}): System performance results of \workname, including average computation, communication, and waiting time (in minutes). }
    \label{fig:exp:real-world-testbed}
\end{figure}

As shown in the left part of Fig.~\ref{fig:exp:real-world-testbed}, we present a comparative analysis of the Bert-Score performance of our proposed \workname~against 5 state-of-the-art baselines on our real-world testbed. The results indicate that \workname~achieves the highest Bert-Score. In particular, \workname~outperforms Qwen, ImageBind-LLM, OneLLM, Language-Bind, and Tarsier by a considerable margin, demonstrating a robust capability in generating accurate captions for low-resolution human behavior understanding tasks. The consistent improvement across different datasets highlights the generalizability and practical applicability of \workname~in real-world scenarios.


\subsubsection{System performance}
\label{sec:exp:system_performance}We evaluate the system performance of \workname, focusing on average computation time, communication time, and waiting time. The waiting time represents the duration the fastest clients must wait for the slowest clients to complete before aggregation. As shown in the right part of Fig.~\ref{fig:exp:real-world-testbed}, we observe that the balance between computation time and communication time highlights the scalability and practicality of our system, making it well-suited for real-world deployment. However, we also note that waiting time constitutes the most significant portion of the overall system time. This is primarily due to varying communication conditions among clients. For instance, the fast client can achieve speeds of up to 20 Mbps, while the slowest clients can only reach 4 Mbps. Consequently, optimizing waiting time emerges as a promising direction for future work. As for the on-device system performance,  we also measure the memory consumption of training LoRA-based efficient fine-tuning, which originally requires around 17.8 GB of memory—exceeding the capacity of Xavier. To address this, we adopt Q-LoRA~\cite{dettmers2023qlora} to reduce memory usage.

\section{Evaluation on low-resolution datasets} 
\label{sec:evaluation}

\subsection{Datasets and Implementations}
\subsubsection{Datasets description}
\begin{table}[t]
\small
\centering
\caption{ Summary of the testbed and public datasets. }
\label{tab:sample_table}
\begin{tabular}{c|cccc}
\hline
\textbf{Dataset} & \textbf{Modality} & 
\makecell{\textbf{\# of} \\ \textbf{Subjects}} &  \makecell{\textbf{\# of} \\ \textbf{Actions}} & \makecell{\textbf{\# of} \\ \textbf{Labeled Videos}}\\ \hline
UTD-D & Depth & 8 & 27 & 864 \\ 
IM-T & Thermal & 2 & 24 & 786 \\ 
SC-IR & Infrared & 10 & 12 & 120 \\ 
RW-D & Depth & 97 & 16 & 7856 \\ 
\hline
\end{tabular}
\end{table}

We select three datasets as auxiliary datasets for evaluation, including the UTD Depth Dataset \textbf{(UTD-D)}~\cite{chen2015utd}, The  Indoor Motion Thermal Dataset \textbf{(IM-T)}~\cite{ThermalIM2023}, and a self-collected Infrared Dataset \textbf{(SC-IR)}. We also provide a detailed dataset description in Table~\ref{tab:sample_table}.

\begin{figure}
    \centering
    \includegraphics[width=\linewidth]{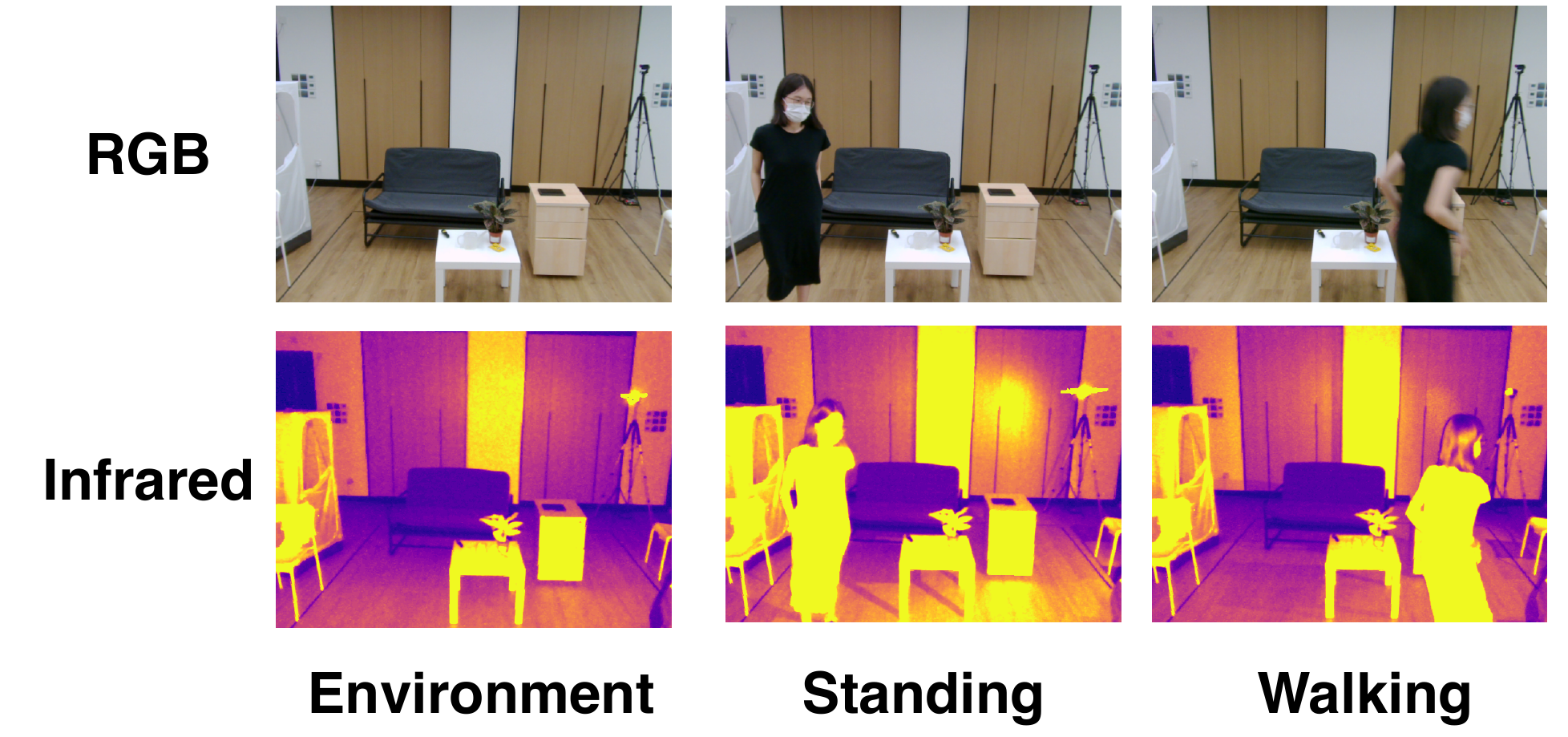} 
    \vspace{-10pt}
    \caption{Illustration example of our self-collected infrared dataset. A significant loss of detailed information is observed compared to the RGB modality.}
    \vspace{-15pt}
    \label{fig:exp:ir}
\end{figure}

\begin{itemize}[leftmargin=*]
    \item \textbf{UTD-D~\cite{chen2015utd}.} The UTD-D dataset is a comprehensive human action dataset captured using a Kinect camera, featuring depth videos of 27 distinct actions performed by 8 subjects. Note that in the UTD-D dataset, the depth videos are recorded and synchronized with the RGB videos.
    \item  \textbf{IM-T~\cite{ThermalIM2023}.} The IM-T dataset contains synchronized RGB-Thermal videos of indoor human motion. It encompasses a wide range of human-object interactions in living room and office settings, with many activities leaving thermal imprints on objects. The dataset consists of 783 video clips. It involves 1 or 2 actors in two different rooms. The remaining clips form a held-out set for generalization testing, featuring either a different actor or a different room.
    \item \textbf{SC-IR.} Since there is a lack of representative infrared datasets specifically designed for HUB systems, we collected data for 12 daily actions performed by 10 subjects to represent common human behaviors (SC-IR) with RGB-Infrared data with 120 video clips in total. These actions include: ``Greeting'', ``Chatting'', ``Sitting'', ``Packing Up'', ``Grooming Oneself'', ``Walking Casually'', ``Taking off Jacket'', ``Sleeping'', ``Chopping Vegetables'', ``Drinking Water'', ``Using Computer'', and ``Sweeping the Floor''. Each volunteer completed all the required actions freely, with the entire process taking approximately 30 minutes per participant. To enhance diversity, each room was configured with varying scenes by adjusting viewing angles or rearranging furniture. Fig.~\ref{fig:exp:ir} provides an example from our SC-IR dataset, showing that infrared data loses significant detail compared to RGB data.
\end{itemize}

\subsubsection{Data caption generation}
Since representative low-resolution vision caption datasets are unavailable, we addressed this limitation by recruiting 3 volunteers to manually caption the entire dataset for our real-world testbed as we mentioned before. For public datasets, we employed Tarsier-34B, to generate captions for the RGB data, as these models are known to perform well on this modality. However, because these models do not accurately capture human behavior, we supplemented their outputs with precise descriptions of actions, forming our final evaluation set.

\subsubsection{Implementation details}
\label{sec:exp:implement_details} For the UTD-D and SC-IR and RW-D dataset, we assign one participant to represent a single client. While in IM-T dataset,  we set up 4 clients  due to the combination of 2 settings and 2 participants with Dirichlet distribution~\cite{jiang2024artfl} with $Dir=1$. In addition, since IM-T is an extremely long-tail dataset, we drop the extremely minority classes to ensure a fair comparison. We use ResNet18 as the base model to train our labeler with the SGD optimizer. The learning rate is set as 0.01 and batch size is 32 for all datasets.
We use Llama-3.1-70b as the LLM for caption generation.
We use LoRA techniques for parameter-efficient fine-tuning to fine-tune Qwen2.5-VL-7B~\cite{Qwen2.5-VL}. Specifically, the learning rate and training epochs were set to $1\times10^{-4}$ and 10, respectively. We set the rank of LoRA as 8 in our experiments.



\subsection{Results on low-resolution datasets}
\label{sec:exp:aux_dataset_results}

\subsubsection{Overall performance.}
\label{sec:exp:performance_on_aux}
\begin{figure*}[t]
    \centering
    \includegraphics[width=\linewidth]{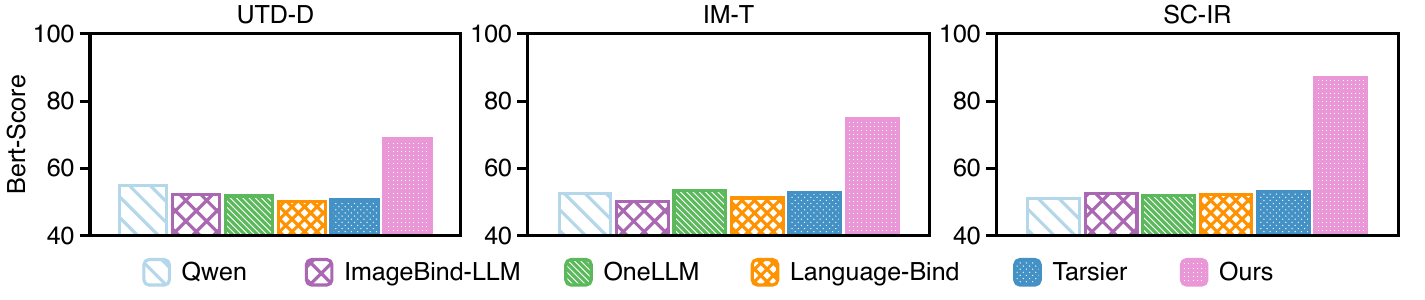} 
    \vspace{-25pt}
    \caption{Performance (Bert-Score) comparison on UTD-D, IM-T, SC-IR datasets. \workname~consistently outperforms the state-of-the-art baselines.}
    \vspace{-10pt}
    \label{fig:exp:1shot-FT}
\end{figure*}

As illustrated in Fig.~\ref{fig:exp:1shot-FT}, we assessed the performance of various models across the UTD-D, IM-T, and SC-IR datasets. The findings demonstrate that \workname~consistently outshines the competing state-of-the-art models on all tested datasets. For example, on the UTD-D dataset, our model recorded a performance score of approximately 75, which is about 10 points higher than its closest competitor, Language-Bind. Similar trends were observed in the IM-T and SC-IR datasets, underscoring the robustness of our approach. These results underscore that \workname~effectively leverages advanced language processing capabilities to enhance the understanding of human behavior in diverse computational tasks.

\begin{figure}[h]
    \centering
    \includegraphics[width=\linewidth]{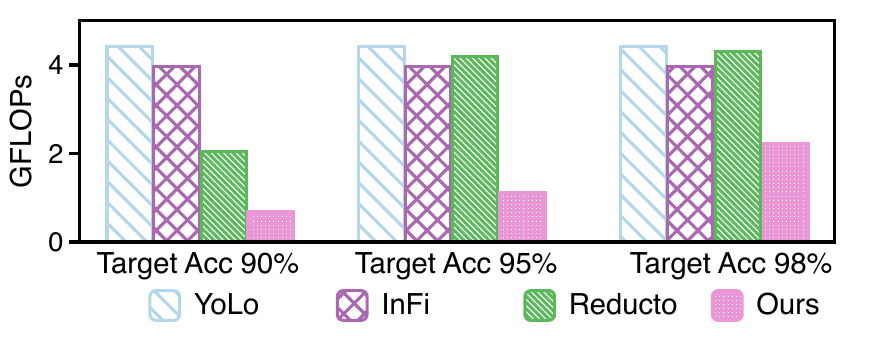} 
    \vspace{-25pt}
    \caption{Results of window-based sensitivity on-camera filtering across three baselines on UTD-D.
    }
    \vspace{-10pt}
    \label{fig:exp:filtering}
\end{figure}

\subsubsection{Performance of window-based sensitivity on-camera filtering}
\label{sec:exp:on-camera-filtering}

In this study, we evaluated the performance of different systems for window-based sensitivity on-camera filtering on IM-T dataset,\footnote{Note that he UTD-D and IM-T datasets do not capture the entry or exit of individuals within the camera's field of view,} targeting varying accuracy levels: 90\%, 95\%, and 98\%. 
We use YOLO~\cite{yolov8_ultralytics} as our baseline and select two strong baselines as follows.
\begin{itemize}
    \item \textbf{Reducto}~\cite{li2020reducto} is an on-camera filtering approach that utilizes cheap vision features to detect the presence of people.
    \item \textbf{InFi}~\cite{yuan2022infi} is a learning-based filtering approach, whose core idea is to train a binary classifier to determine the presence of people.
\end{itemize}
As shown in  Fig.~\ref{fig:exp:filtering}, The results demonstrated that \workname~exhibits superior performance across all predefined accuracy thresholds among YOLO, Reducto, and InFi baselines.\footnote{Note that while InFi requires training a binary classifier, in our experiments we only consider the inference cost. We use ResNet-8 as the base model for InFi.} This model is particularly effective in reducing computational costs while maintaining comparable accuracy levels. This analysis emphasizes the efficacy of our approach in applications that demand high precision and sensitivity in real-time video processing environments.


\begin{figure}[h]
    \centering
    \includegraphics[width=\linewidth]{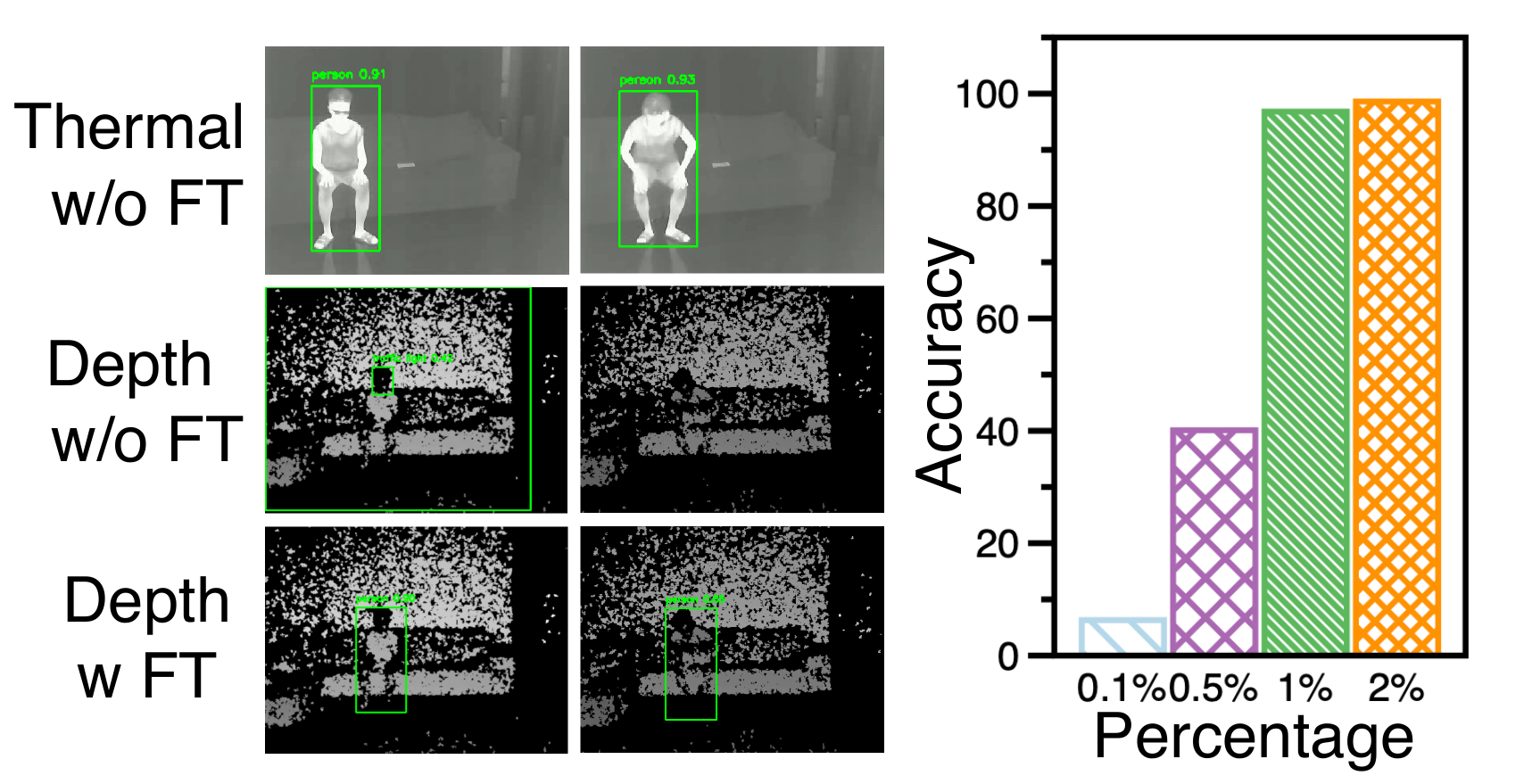}
    \vspace{-15pt}
    \caption{ (\textit{Left}): Visualization of human detection results on low-resolution data of varying quality using pretrained and fine-tuned YOLO. \texttt{FT} stands for \texttt{Fine-tuning}. (\textit{Right}): The effect of using different percentages of data for fine-tuning.
    }
    \label{fig:yolo_results}
\end{figure}

\subsubsection{Performance of fine-grained human action capturing}
\label{sec:exp:vis_hac}
From our experience, we can achieve reliable human detection with pretrained YOLO on low-resolution vision data. As shown in the left part of Fig.~\ref{fig:yolo_results}, for high-quality data, such as the thermal data shown in the first row, the pretrained YOLO model performs effectively without additional fine-tuning. However, for lower-quality data, such as the depth data shown in the second row, YOLO struggles to detect objects accurately. However, after fine-tuning YOLO on a small dataset, it significantly improves detection accuracy, as demonstrated in the third row.  Moreover, in the right part of Fig.~\ref{fig:yolo_results}, we notice that we only use $1\%$ label data for fine-tuning the YOLO models to significantly improve detection accuracy, achieving $96\%$ performance.

\subsubsection{Performance of Top-K action selection}
\label{sec:exp:top-k}
\begin{figure}[t]
    \centering
    \includegraphics[width=\linewidth]{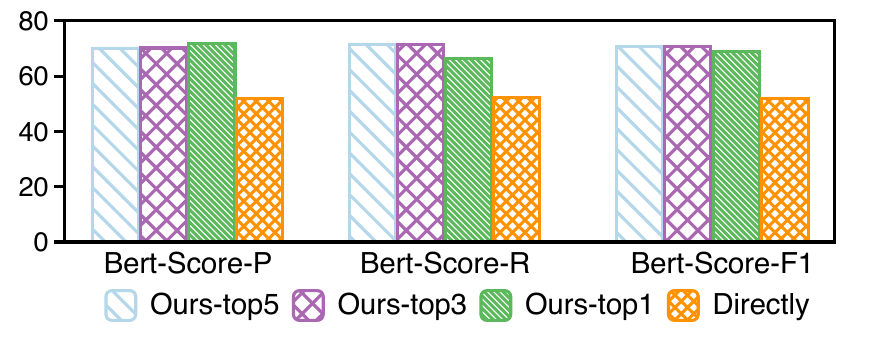}
    \vspace{-20pt}
    \caption{Results of Top-K action selection on UTD-D Dataset. We select top-1, top-3, top-5 as our final input. ``Directly'' indicates that we use the raw time-series sequence as input. }
    \vspace{-10pt}
    \label{fig:top-k}
\end{figure}

To evaluate our captioner in \S\ref{sec:method:stage2}, we conduct an experiment to assess the effect of top-K action selection on the UTD-D dataset. As shown in Fig.~\ref{fig:top-k}, we present the Bert-Score metrics, including Precision, Recall, and F1-score, for three variations: top-5, top-3, and top-1 actions. The results show that top-5 (Ours-top5) and top-3 (Ours-top3) selections outperform top-1 selection, suggesting that using multiple action predictions provides the most balanced and effective results in terms of precision and recall. The primary reason is that including more actions provides richer information.

\subsubsection{Effect of the quality of labeler}
\label{sec:exp:effect_labeler}
\begin{figure}[t]
    \centering
    \includegraphics[width=\linewidth]{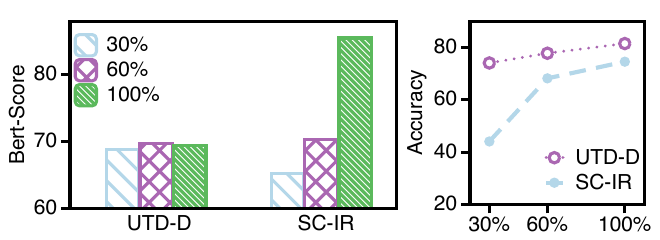}
    \vspace{-15pt}
    \caption{ (\textit{Left}): Bert-Score across varying amounts of labeled data. (\textit{Right}):  Accuracy of Labeler across varying amounts of labeled data. }
    \vspace{-10pt}
    \label{fig:exp:labeled_amount}
\end{figure}

We evaluate the impact of labeler quality in \workname~by training the labeler with varying amounts of labeled data. In particular, we examine the impact of unlabeled data on the performance of UTD-D and SC-IR datasets under different proportions of labeled data (30\%, 60\%, and 100\%). As shown in the left part of Fig.\ref{fig:exp:labeled_amount}, we observe that for the UTD-D dataset, there are only marginal improvements as more labeled data is added, while SC-IR exhibits a steep upward trajectory as labeled data increases. The primary reason lies in the quality of the labeler. As shown in the right part of Fig.~\ref{fig:exp:labeled_amount}, UTD-D, being relatively clean, requires limited labeled data for a well-trained labeler, while SC-IR benefits significantly from more labeled data. This highlights that the high-quality labeler can provide better captions, thus improving the overall performance of \workname.

%% file: 2_related.tex
\section{Related Work}
\label{sec:related_work}
%
\subsubsection*{\textbf{Label-based Vision HBU}}
Label-based vision approaches for HBU can be broadly categorized based on the modality into high-resolution and low-resolution data. 
High-resolution vision data, i.e., RGB videos, provide detailed information about subjects, making them particularly effective for understanding fine-grained activity~\cite{siddiqui2024dvanet,zhou2023unified,reilly2024just}.
Recent studies also proposed the multi-label action recognition tasks~\cite{zhang2021multi}.
However, using multi-label words to describe human behavior is inferior to detailed captions due to limited semantic details and human-computer interaction experience.
In addition, high-resolution modalities pose significant privacy concerns due to the inherently detailed and sensitive nature of the data.


Low-resolution modalities, such as depth, infrared, and thermal imaging, offer a promising solution to mitigate significant privacy concerns~\cite{sun2022human}.
Prior works~\cite{sanchez20223dfcnn,sanchez2020exploiting} have tailored neural network architectures to leverage the unique characteristics of depth data, enhancing the extraction of spatiotemporal features from depth videos for activity recognition. 
Studies~\cite{ouyang2022cosmo,xu2023mesen} also explored improving features extracted from depth data using complementary modalities during training, resulting in robust recognition performance. 
For infrared and thermal data, works such as~\cite{dai2021multi,mehta2021motion,hou2022low} specially design network architectures to extract discriminative features from low-resolution infrared or thermal images for activity recognition and tracking. 
However, the aforementioned approaches use a single label to describe the action, which cannot precisely capture the details of the action and provide information about the surrounding environment.




\subsubsection*{\textbf{LLM-Driven HBU}}
Previous studies have explored various methods to enable LLMs to understand human behavior~\cite{xu2024penetrative,ji2024hargpt}.
They directly feed the raw sensor data and incorporate expert knowledge about sensor waveforms into the prompt to enhance LLMs' understanding of sensor data and improve their performance in HBU tasks.
Some recent studies~\cite{chen2024towards,ouyang2024llmsense,yang2025socialmind} also explore leveraging LLMs for high-level summarization and reasoning on sensor data events, rather than focusing on understanding raw sensor data, which is used in many HBU applications.
Some studies have also explored projecting multimodal sensor data into the embedding space of LLMs~\cite{xiong2024novel}.
SensorLLM~\cite{li2024sensorllm} fine-tune pre-trained foundation models on HAR tasks using various adapter architectures.
Weng et al.~\cite{weng2024large} propose to utilize the knowledge of existing Vision-FMs to improve the model performance on Radio-Frequency (RF)-based HAR tasks using few-shot learning. 
Xiong et al.~\cite{xiong2024novel} have designed a two-stage framework that enables LLMs to understand sensor data for HAR tasks. 
PeVL \cite{zhang2024pevl} utilizes contrastive learning to align vision, text, and pose modalities to achieve fine-grained HAR. However, the aforementioned works are not designed to handle low-resolution data.

\subsubsection*{\textbf{Understanding Low-Resolution Vision Data via LVLMs}}
 
Although high-resolution visual data, such as RGB images, has been widely incorporated into multi-modality LVLMs, there remains a gap in understanding low-resolution modalities within these models.
ImageBind~\cite{girdhar2023imagebind} and LanguageBind~\cite{zhu2023languagebind} bind different modalities into pairs for alignment, allowing low-resolution data like depth to align with RGB and text modalities and performing many zero-shot tasks.
Meta-Transformer~\cite{zhang2023meta} utilized a unified framework to learn from heterogeneous modalities, despite the absence of data pairing. 
Studies like OneLLM~\cite{han2024onellm} further project depth data into the token space of LLMs.
PathWeave~\cite{yu2024llms} took advantage of continuous learning and incremental training to improve scalability to new modalities, including depth. However, to the best of our knowledge, \workname~is the first system which is designed for low-resolution vision data.

%% file: 8_con.tex
\section{Discussions}
\label{sec:discussion}
\subsubsection*{\textbf{Scalability to heterogeneous devices and distribution}}
\workname~is designed to be highly adaptable, requiring minimal calibration, re-training, or fine-tuning when deployed across heterogeneous devices or environments. However, when utilizing low-memory, low-computation devices such as the NVIDIA Jetson Xavier Nano or TX2, it becomes necessary to apply other quantization techniques or select lower-rank configurations of LoRA to ensure efficient operation. In addition, our real-world testbed demonstrates that the waiting time constitutes the most significant delay in \workname, echoing in \S\ref{sec:exp:system_performance}. Developing strategies to optimize and reduce this delay will be an important direction for future research and system improvement.


\subsubsection*{\textbf{Scalability to multiple people scenarios}} 
\workname~reduces effectiveness in scenarios involving multiple people performing multiple actions simultaneously. This constraint arises from the current limitations of our labeler and captioner, which are designed for single-person scenarios since it is challenging to naturally encode multi-person information using pseudo labels from both temporal and spatial perspectives.  As a result, \workname~is better suited for environments where individual actions need to be recognized, such as personalized healthcare monitoring or single-user assistance systems. However, this limitation does not undermine the system’s value, as single-person scenarios are highly relevant in many practical applications. 


\subsubsection*{\textbf{Scalability to more modalities, more sensor data and open-set learning.}}
\workname~currently only focuses on low-resolution vision data including depth, thermal, and infrared. However, the key insight of \workname, leveraging class-aware guidance for precision caption generation, can also be scalable to other modalities such as IMU and point cloud.  
In addition, other data augmentation approaches, such as GAN~\cite{jiang2022pgada}, diffusion models~\cite{croitoru2023diffusion}, and 3D-GS~\cite{chen2024survey}, can also be considered to further improve the performance of \workname. \workname~enables LVLM fine-tuning and facilitating human behavior understanding. However, our current evaluation dataset is limited in size, constraining the capabilities of the trained LVLMs.  We envision that our pipeline can be scaled to accommodate more data scenarios, further enhancing the generalizability and open-set performance of LVLMs~\cite{Qwen2.5-VL,yang2023edgefm}.





\section{Conclusion}
\label{sec:conclusion}
In this paper, we propose \workname, a system tackling low-resolution HBU in heterogeneous vision systems through contrastive-oriented data labeler, physical-knowledge-guided captioner, and loRA-based efficient fine-tuning. By adapting LVLMs to low-resolution data, \workname~improves HBU performance while remaining efficient. Evaluations on real-world datasets show its effectiveness for on-device HBU, thus it can further enhance applications like healthcare monitoring and personal assistants.